\documentclass[letterpaper, 10 pt, conference]{ieeeconf}  

\IEEEoverridecommandlockouts                              

\usepackage{graphics} 
\usepackage{xcolor}
\usepackage{epsfig} 
\usepackage{mathptmx} 
\usepackage{times} 
\usepackage{amsmath} 
\usepackage{amssymb}  
\usepackage{graphicx} 

\usepackage{algorithm}
\usepackage[ruled,vlined,algo2e]{algorithm2e}

\usepackage{hyperref}

\usepackage{theoremref} 
\usepackage{thmtools}
\usepackage{thm-restate} 
\usepackage{amsxtra} 

\usepackage{textcase}

\newcommand{\gaifo}{\textsc{gai}f\textsc{o}}

\newcommand{\ifo}{\textsc{i}f\textsc{o}}
\newcommand{\voila}{\textsc{voila}}

\newcommand{\rae}{\textsc{rae}}

\usepackage{subcaption} 
\usepackage{cite} 
\usepackage{xcolor}
\let\labelindent\relax
\usepackage{enumitem}

\title{\Large \bf
VOILA: Visual-Observation-Only Imitation Learning for Autonomous Navigation
}

\author{Haresh Karnan$^{1}$, Garrett Warnell$^{2,3}$, Xuesu Xiao$^{3}$ and Peter Stone$^{3,4}$ 
\thanks{$^{1}$ The University of Texas at Austin, Department of Mechanical Engineering {\tt\footnotesize haresh.miriyala@utexas.edu}}%
\thanks{$^{2}$ Army Research Laboratory {\tt\footnotesize garrett.a.warnell.civ@mail.mil}}%
\thanks{$^{3}$ The University of Texas at Austin, Department of Computer Science {\tt\footnotesize xiao@cs.utexas.edu}}%
\thanks{$^{4}$ Sony AI {\tt\footnotesize pstone@cs.utexas.edu}}%
}

\begin{document}

\maketitle
\thispagestyle{empty}
\pagestyle{empty}

\begin{abstract}

While imitation learning for vision-based autonomous mobile robot navigation has recently received a great deal of attention in the research community, existing approaches typically require state-action demonstrations that were gathered using the deployment platform.
However, what if one cannot easily outfit their platform to record these demonstration signals or---worse yet---the demonstrator does not have access to the platform at all?
Is imitation learning for vision-based autonomous navigation even possible in such scenarios?
In this work, we hypothesize that the answer is yes and that recent ideas from the Imitation from Observation (\ifo{}) literature can be brought to bear such that a robot can learn to navigate using only ego-centric video collected by a demonstrator, even in the presence of viewpoint mismatch.
To this end, we introduce a new algorithm, Visual-Observation-only Imitation Learning for Autonomous navigation (\textsc{voila}), that can successfully learn navigation policies from a single video demonstration collected from a physically different agent.
We evaluate \textsc{voila} in the AirSim simulator and show that \textsc{voila} not only successfully imitates the expert, but that it also learns navigation policies that can generalize to novel environments.
Further, we demonstrate the effectiveness of \textsc{voila} in a real-world setting by showing that it allows a wheeled Jackal robot to successfully imitate a human walking in an environment while recording video with a handheld mobile phone camera.

\end{abstract}


\begin{figure*}[!tb]
    \centering
    \includegraphics[scale=0.45]{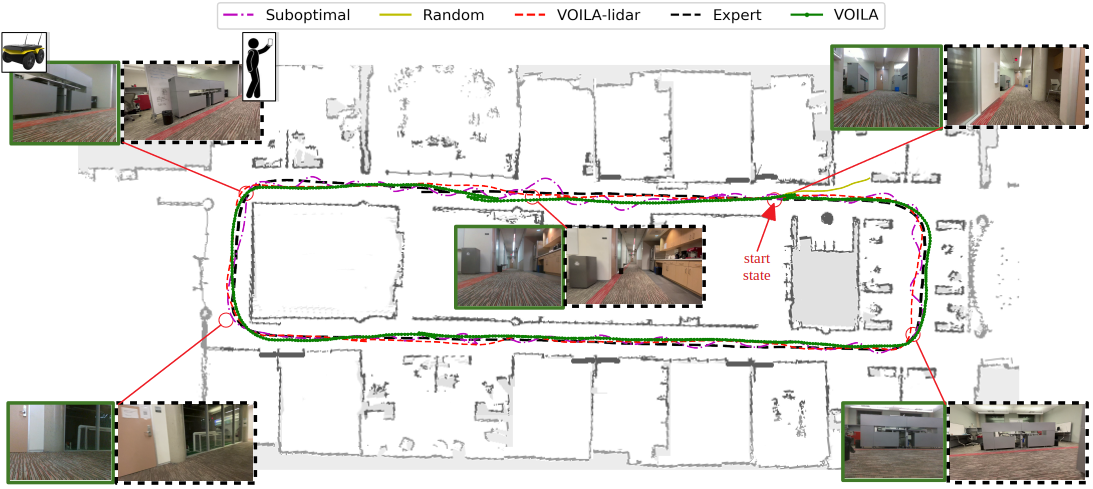}
    \caption{Policy rollout trajectories of the \voila{} agent (green) successfully imitating a demonstration behavior (black) of patrolling a rectangular hallway clockwise. The demonstration consists of a video gathered by a human walking while using a handheld camera that is considerably higher than the robot's camera (introducing significant viewpoint mismatch). We see that the \voila{} agent is able to successfully imitate the expert demonstration even in the presence of this egocentric viewpoint mismatch.}
    \label{fig:voila_trajs_w_mismatch}
\end{figure*}

\section{INTRODUCTION}

Enabling vision-based autonomous robot navigation has recently been a topic of great interest in the robotics and machine learning community \cite{ bojarski, cil, byronboots}.
Imitation learning in particular has emerged as a useful paradigm for designing vision-based navigation controllers. Using this paradigm, the desired navigation behavior is first demonstrated by another agent (usually a human), and then a recording of that behavior is supplied as training data to a machine learner that tries to find a control policy that can mimic the demonstration.
To date, most approaches in the navigation domain that use imitation learning require demonstration recordings that contain both state observations (e.g., images) and actions (e.g., steering wheel angle or acceleration) gathered onboard the deployment platform \cite{ bojarski, cil, offroadoa,  end2endsurvey}.

While these existing imitation learning approaches have proved successful in certain scenarios, there are situations in which it would be beneficial to relax the requirements they impose on the demonstration data.
For example, if we wish to collect a large number of demonstrations from many experts, it may prove too difficult or costly to arrange for each expert to operate specific deployment platforms, which are often expensive or difficult to transport. Additionally, it might be costly to outfit all demonstration platforms with instrumentation to record the control signals with the demonstration data.
However, due to the low cost and portability of video cameras, it may still be feasible to have demonstrators record ego-centric video demonstrations of their navigation behaviors while operating a different platform. 
Demonstrations of this nature would consist of video observations only (i.e., they would not contain control signals), and, because of the difference in platform, the videos would likely exhibit ego-centric viewpoint mismatch compared to those that would be captured by the deployment platform.
One example of such data is the plethora of vehicle dashcam videos available in publicly-accessible databases \cite{dashcamvids} or on YouTube.
Another example is video demonstrations of robot behaviors generated by proprietary code that one would like to mimic on the same or different robot hardware. Unfortunately, to the best of our knowledge, there exist no current imitation learning techniques for vision-based navigation that can leverage such demonstration data.

Fortunately, recent work in \textit{Imitation from Observation} (\ifo{}) \cite{farazijcaisurvey}---imitation learning in the absence of demonstrator actions---has shown a great deal of success for several related tasks. For example, work in this area has been able to learn from video-only demonstrations for both simulated and real limbed robots \cite{farazridm, gaifopaper, TCN, zeroshotil}.
However, no literature of which we are aware has considered whether these \ifo{} techniques can be applied to the vision-based autonomous navigation problem we have outlined above. This problem is especially challenging since physical differences in the demonstration platform introduce viewpoint mismatch in the video demonstrations.

In this paper, we hypothesize that it is possible to perform imitation learning for vision-based autonomous navigation using video-only demonstrations collected using a physically different platform.
To this end, we introduce a new \ifo{} technique for vision-based autonomous navigation called {\em Visual-Observation-only Imitation Learning for Autonomous navigation} (\voila{}).\footnote{A preliminary version of this work was presented at the 2021 AAAI Spring Symposium on Machine Learning for Navigation.} 
To overcome viewpoint mismatch, \textsc{voila} uses a novel reward function that relies on off-the-shelf keypoint detection algorithms that are themselves designed to be robust to egocentric viewpoint mismatch. This novel reward function is utilized to drive a reinforcement learning procedure that results in navigation policies that imitate the demonstrator.

We experimentally confirm our hypothesis both in simulation and on a physical Clearpath Jackal robot. We compare \voila{} against a state-of-the-art \ifo{} algorithm \gaifo{} \cite{gaifopaper}, and show that \voila{} can learn to imitate an expert's visual demonstration in the presence of viewpoint mismatch while also generalizing to environments not seen during training. Additionally, we demonstrate the flexibility of \voila{} by showing that it can also support vision-based training of navigation policies with inputs other than camera images.

\section{Background and Related Work}

The proposed approach, \textsc{voila}, performs reinforcement learning (RL) using a novel reward function based on image keypoints in order to accomplish imitation from observation for autonomous navigation with viewpoint mismatch. In this section, we review related work in autonomous robot navigation, imitation from observation, and in computer vision techniques for visual feature extraction.

\subsection{Machine Learning for Autonomous Navigation}

The use of machine learning methods in the design of autonomous navigation systems goes back several decades, though recent years have seen a spike in interest from the research community \cite{navigationbyimitation, cil, byronboots, xuesusurvey, semanticviz}.
One of the earliest successes was reported by Pomerleau \cite{alvinn}, in which a system called \textsc{alvinn} used imitation learning to train an artificial neural network that could perform lane keeping based on demonstration data generated in simulation.
Since then, several improvements, both in the amount and type of demonstration data and in network architecture and training, have been proposed in the literature.
In particular, LeCun {\em et al.} proposed the use of a convolutional neural network (CNN) to better process real demonstration images for an off-road driving task \cite{offroadoa}, and, more recently, Bojarski {\em et al.} reported that gathering a large amount of real-world human driving demonstration data and applying data augmentation made it possible to train even more-complex CNN architectures to perform lane keeping \cite{bojarski}.

While the aforementioned approaches each use end-to-end imitation learning to find autonomous navigation policies, other machine learning for autonomous navigation work has adopted the alternative training paradigm of RL.
Chang {\em et al.} propose using off-policy Q-learning from video demonstration data to learn goal conditioned hierarchical policies for semantic navigation \cite{semanticviz}. While their approach also uses video demonstrations with no action labels, they learn a goal conditioned policy with access to thousands of navigation video examples whereas in our work, the focus is on imitating an expert's video demonstration in the presence of viewpoint mismatch using a single video-only demonstration. Gupta {\em et al.} propose a context translation network to imitate an expert demonstration in the presence of viewpoint mismatch. However, their approach requires multiple demonstrations with differing camera viewpoints in each demonstration to provide context signals. Additionally, their approach only deals with third-person viewpoint mismatch and does not consider the egocentric viewpoint mismatch problem that is considered in this work. 

The work closest to ours is that of Kendall {\em et al.} \cite{driveinaday}, in which the proposed system learns a navigation policy using RL, where the reward function is the total distance travelled by their autonomous vehicle before a human driver intervenes (to, e.g., prevent collisions).
However, unlike \textsc{voila}, Kendall {\em et al.} utilize experience gathered exclusively by the learning platform itself, considering neither imitation from observation nor the particular problem of viewpoint mismatch. 

\subsection{Imitation from Observation}

Recently, there have been a number of imitation from observation (\ifo{}) techniques introduced in the literature, including the adversarial approach, \gaifo{}, proposed by Torabi {\em et al.} \cite{gaifopaper, farazproprioception} which we use as a comparison point in this paper. 
In \gaifo{}, the reward signal is provided by a learned discriminator network which seeks to reward state transitions similar to those present in the demonstration and penalize --- if it can tell the difference --- state transitions that come from the imitator.
While \textsc{gaifo} has been shown to be successful in both low- and high-dimensional observation spaces, it has thus far only been applied to continuous control tasks for limbed agents.
Moreover, as we will show in our experiments, while \gaifo{} is able to imitate the expert when the egocentric viewpoints between the expert and the imitator match, it is unable to do so in the presence of viewpoint mismatch.

Viewpoint mismatch in \ifo{} has been previously considered in the work by Sermanet {\em et al.} \cite{TCN}, which proposes Time Contrastive Networks (\textsc{tcn}s). \textsc{tcn}s use a triplet loss metric to learn a feature space embedding which is then used for rewarding the agent to imitate the expert. While both \voila{} and \textsc{tcn}s are robust to viewpoint mismatch, \textsc{tcn}s require demonstration data with multiple viewpoints in the same timestep in order to learn an embedding space that is robust to viewpoint mismatch, whereas \voila{} achieves this robustness by leveraging image feature detection algorithms (e.g \textsc{sift} \cite{sift}) commonly used in \textsc{slam} that are themselves designed to be robust to viewpoint mismatch.

\subsection{Feature Detection and Matching}
To overcome viewpoint mismatch, \voila{} utilizes a novel reward function that relies on local image features such as keypoints and their descriptors.
Keypoints have been used for decades to solve challenging tasks such as image verification, matching and retrieval.
More recently, deep-learning-based keypoint extractors such as \textsc{superpoint} \cite{superpoint} have been shown to be more successful than classical approaches.
In this work, we use \textsc{superpoint} to detect keypoints and descriptors, and we determine keypoint matches between two images using the typical method based on the two nearest neighbors in descriptor space \cite{superpoint}. However, in principle \voila{} can be used with any local feature detector or feature matching algorithm.
Several works have proposed learning a keypoint detector specific to the imitation learning task \cite{mbrlkeypoint, keypointsintofuture}. Unlike such approaches, \voila{} uses an off-the-shelf keypoint extractor that is not trained specifically for the navigation task. 

\section{\voila{}}
In this section, we formulate the imitation learning problem for the task of autonomous visual navigation, which we pose as a reinforcement learning problem with a demonstration-dependent reward.
The critical contribution of \voila{} is the development of this particular reward function, which we describe in detail below.

\subsection{Preliminaries}
We treat autonomous visual navigation as a RL problem where the environment is a Markov decision process. 
At every time step $t$, the state of the agent is described by $s_t\in\mathcal{S}$, the observation of the agent is described by $O_t\in\mathcal{O}$, and an action $a_t \in \mathcal{A}$ is sampled from the agent's policy $a_t \sim \pi(\cdot|O_t)$.\footnote{While \voila{}'s reward function depends on camera images, the imitation policy can actually be learned over {\em any} appropriate state representation---vision-based or otherwise. We explore this further in Section \ref{sec:experiments}.}
A single expert demonstration is represented as a set of $n$ sequential observations $\mathcal{D}^e = \{O^e_1, O^e_2, \dots ,O^e_n\}$. Performing this action in the environment leads to a next state $s_{t+1} \sim T(\cdot|s_t,a_t)$, where $T$ is the unknown transition dynamics of the agent in the environment. For this specific transition, the agent receives a reward, $r_{t+1} \in \mathbb{R}$, which is a function of both the agent's transition tuple and the demonstration, i.e., $r_{t+1} = R(O_t, a_t, O_{t+1}; \mathcal{D}^e)$. The relative utility of near-term and long-term reward is controlled using the discount factor $\gamma \in (0, 1]$.
The RL objective is to find a policy $\pi$ that maximizes the expected sum of discounted rewards $\mathbb{E}[\Sigma_{t=0}^{\infty} \gamma^t R(O_t, a_t, O_{t+1}; \mathcal{D}^e)]$. 

\subsection{Reward Formulation for Imitation Learning}


For each transition $(O_t, a_t, O_{t+1})$ experienced by the learner, we require a reward $r_{t+1}$ such that the learner, by optimizing the RL objective with this reward, can learn to imitate the demonstration.
In particular, because we wish to perform learning in real time, we seek a dense reward function that provides feedback at each timestep without delay.
Since the expert demonstrations are from a physically different agent, there can be significant ego-centric viewpoint mismatch between the observation spaces of the learner and the demonstrator as shown in Fig. \ref{fig:voila_trajs_w_mismatch}. Such a mismatch poses a challenge to designing a good reward function since it is not immediately clear how to compare images from different viewpoints. Hence, we introduce here a novel reward function based on keypoint feature matches between the expert and the imitator's ego-centric observations for the task of imitation learning for visual navigation. Keypoint detectors have been extensively used in the computer vision community for several decades to solve challenging tasks like structure-from-motion (\textsc{s}f\textsc{m}), visual \textsc{slam} and hierarchical localization. Recent keypoint detection algorithms like \textsc{superpoint} \cite{superpoint} provide invariance to perspective distortion, scaling, translation, rotation, viewpoint mismatches, and varied lighting conditions between the key and query images. Hence, we use keypoint detectors to help define the reward function to learn visual navigation policies from demonstrations provided by any other agent. 

The reward function we propose relies on a quantity that we call {\em match density}. We define the match density $d(O_1, O_2)$ between two images $O_1$ and $O_2$ as the ratio of the number of keypoint matches between $O_1$ and $O_2$, and the total number of detected keypoints in $O_2$.
$d(O_1, O_2) \in [0, 1]$, assuming there is always a non-zero number of keypoints detected in an image.
Additionally, instead of imposing a temporal alignment constraint, we define the reward for a particular transition by searching the demonstration for the image which is most visually similar to the learner's current observation.
Here, we define the most visually similar image in the expert demonstration to be the one that has the highest match density with $O_t$, which we denote as $O^e_{i_t}$.
For convenience of notation, we denote $O^e_{i_t+1}$ (the next image after $O^e_{i_t}$ in the demonstration sequence), as $\hat{O}^e_{i_t}$. We also point out here that, while $\hat{O}^e_{i_t}$ is the image in the demonstration sequence that follows $O^e_{i_t}$, it may differ from the image in the demonstration sequence that is most similar to $O_{t+1}$ (i.e., $O^e_{i_{t+1}}$).

Using the concepts described above, we now define the proposed reward function for \textsc{voila}:

\begin{equation}
   R(O_t, a_t, O_{t+1}; \mathcal{D}^e) = 
   \begin{cases} 
      F + V - \lambda ||a^{steer}_t|| &, alive \\
      -10 &,  done \\
   \end{cases} \; ,
  \label{rewardfunc}
\end{equation}
where $F=d(O_{t+1}, O^e_{i_{t+1}})$ and $V=\gamma * d(O_{t+1}, \hat{O}^e_{i_t}) - d(O_t, \hat{O}^e_{i_t})$. 
If the robot is in the {\em done} state, i.e., it has crashed (as detected in AirSim, or by the trainer in physical experiments) or the number of keypoint matches drop below 10, the agent receives a penalty reward of $-10$.
Otherwise, the agent is in the $alive$ state, and we assign a reward that depends on terms $F$ and $V$.
The $F$ term assigns reward value based on the match density encountered at the next observation $O_{t+1}$ that the agent ends up in the transition.
This component encourages the agent to stay on the demonstrated trajectory.
The $V$ term is similar to a potential-based shaping term, and rewards a transition based on the difference in the match densities with the next expert observation $\hat{O}^e_{i_t}$ and the imitator's observations.
This component encourages the imitator to find a policy that exhibits similar state transitions to those experienced by the expert.
We additionally found that adding the action penalty term with a $\lambda$ of $0.01$ penalizes the agent for making large steering changes. The expert image retrieval step is performed in real-time using feature matching, and is outlined in the implementation section.

\section{Implementation}
In this section, we provide specific implementation details of \voila{} including those related to representation learning, keypoint feature extraction, and the network architectures.

\subsection{Representation Learning}

Representation learning using unsupervised learning is a powerful tool to improve the sample efficiency of deep RL algorithms. Instead of learning a navigation policy over high dimensional image space, \textsc{voila} uses a latent representation of the image and learns the navigation policy over this latent code as input to the policy. Specifically, \textsc{voila} uses a Regularized Auto Encoder (\textsc{rae}) \cite{rae} to learn a latent posterior of the visual observations of the imitator.
The imitating control policy is then learned using RL with the latent code $z_t = g_\phi(O_t)$ as the input to the policy network, where $g_\phi$ is the encoder of the \textsc{rae} with weights $\phi$.
A ResNet-18 encoder-decoder network architecture is used for the \textsc{rae} and is trained for the task of image reconstruction, with data collected from the imitator using random rollouts.
The input images are of size $256 \times 256$ and the size of the latent dimension is $512$.
Random cropping and random affine image augmentations are utilized to regularize training.

\subsection{Keypoint Feature Extraction}
As a preprocessing step, all \textsc{superpoint} features detected from expert observations are stored in a buffer. At the start of an episode (for the first frame), the nearest expert observation $O^e_{i_t}$ to $O_t$ is retrieved by linearly searching for the closest expert observation in $\mathcal{D}^e$ with the maximum feature matches. As the episode unfolds, instead of exhaustively searching for the closest expert image at every transition, the search is restricted over the three next expert observations forward in time from the previous closest expert image in $\mathcal{D}^e$. At each transition, \textsc{superpoint} keypoints and descriptors are extracted and used to retrieve the closest expert image and compute the reward according to Equation \ref{rewardfunc}. Note that the \textsc{superpoint} keypoint descriptors extracted in an image are the local features and not the global features for that image. Hence, we explicitly train an \rae{} to learn a compressed global representation of the image for training the navigation policy, as described in the previous section. 

\subsection{Navigation Policy Architecture}
\label{experimental_setup}

We model the navigation policy using a 3-layer, fully-connected neural network, with 256 neurons per layer. We perform frame-stacking with two consecutive latent codes of observations in time to alleviate effects of partial observability in the environment. We also additionally include the most recent action performed by the agent as a part of the state. We use Soft Actor-Critic (\textsc{sac}) \cite{sac}, an off-policy RL algorithm, to learn the navigation policy $\pi$.

\begin{figure}[!tb]
    \centering
    \includegraphics[scale=0.18]{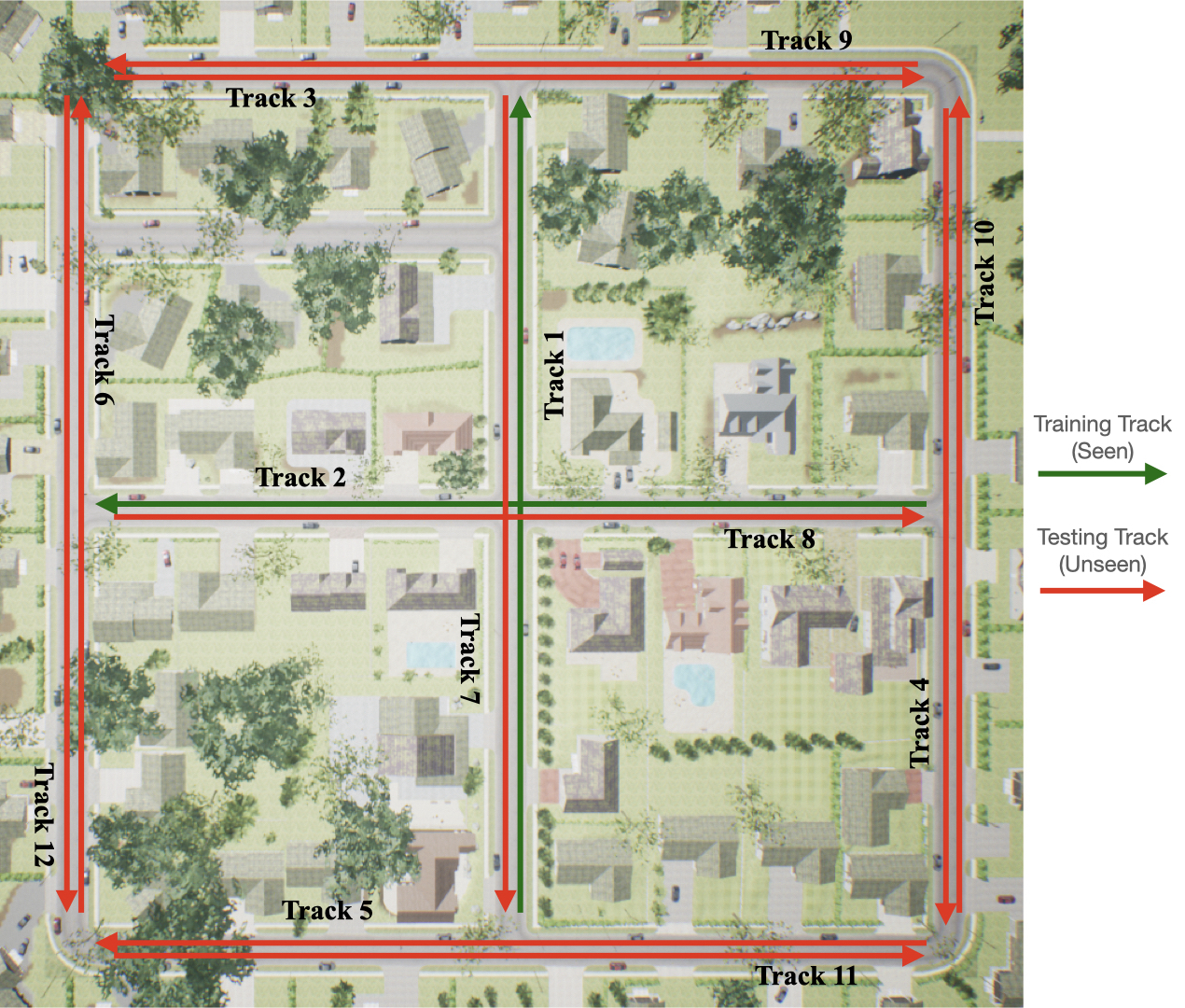}
    \caption{Aerial image of the AirSim simulation environment. Green lines show the tracks used to train the agent and red lines show the tracks unseen by the agent.}
    \label{fig:airsim_top}
\end{figure}

\begin{figure*}[!tb]
\begin{subfigure}{\columnwidth}
    \centering
    \includegraphics[scale=0.45]{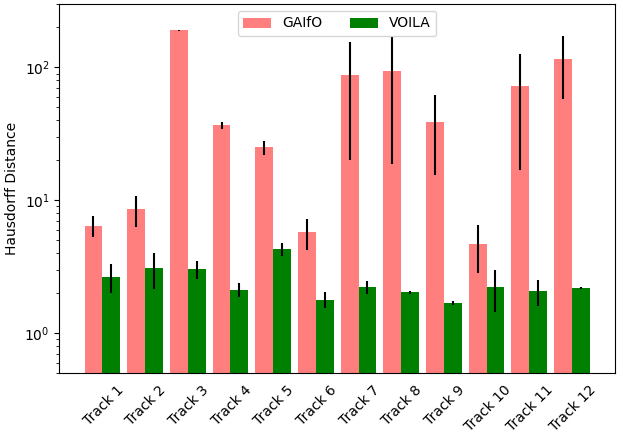}
    \caption{Without viewpoint mismatch between expert and imitator}

    \label{fig:bar_wo_vp}
\end{subfigure}
\begin{subfigure}{\columnwidth}
    \centering
    \includegraphics[scale=0.45]{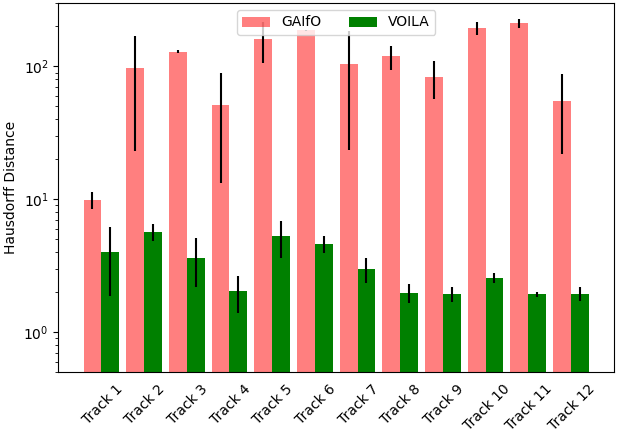}
    \caption{With viewpoint mismatch between expert and imitator}
    \label{fig:bar_w_vp}
\end{subfigure}
\label{fig:bargraphs}
\caption{Imitation performance of policies learned using \voila{} and \gaifo{} in AirSim. The y-axis shows Hausdorff distance between expert and imitator's trajectories, averaged across five trials (lower distance indicates behavior more similar to the expert). We see that with viewpoint mismatch, the \gaifo{} agent is unable to imitate the expert, whereas \voila{} is unaffected by viewpoint mismatch and results in policies that induce behavior closer to that of the demonstrator. Tracks 1 and 2 were used for training, and other tracks were unseen by the agent while learning.}
\end{figure*}

\section{Experiments}
\label{sec:experiments}

We now describe the experiments that we performed to evaluate \voila{}.
The experiments are designed to answer the following questions:
\begin{enumerate}[label=($Q_\arabic*$)]
    \item Is \voila{} capable of learning imitative policies from video demonstrations that exhibit viewpoint mismatch?
    \item How well do policies learned using \voila{} generalize to environments unseen during training?
\end{enumerate}

To answer the questions above, we perform experiments using both a simulated autonomous vehicle and a real Clearpath Jackal robot. In the simulation and real-world environments, the \textsc{voila} agent is tasked with imitating ``road following" and ``hallway patrol" tasks, respectively. The reactive visual navigation policy learned using \voila{} can be classified as moving-goal navigation, i.e., no goals are used or planning performed as in classical waypoint-driven navigation \cite{xuesusurvey}. The objective of the \voila{} agent is to imitate the expert's visual demonstration by learning an end-to-end navigation policy, even in the presence of viewpoint mismatch. To quantify performance of imitation policies, we compute the Hausdorff distance metric (lower is better) between trajectories generated on a held-out set of environments.

\subsection{Simulation Experiments in AirSim}
In our simulation experiments, we answer questions $Q_1$ and $Q_2$ using the outdoor `Neighborhood' environment in AirSim \cite{airsim} and learn the task of road following, i.e., driving on a straight road while avoiding collisions with obstacles such as parked cars along the curb. In learning to perform this task, the agent must also contend with varied lightning conditions and the presence of shadows and road intersections along its path. To this end, we pick 12 straight road segments (tracks) in the AirSim environment, shown in Fig. \ref{fig:airsim_top}. Tracks 1 and 2 marked in green are used for training the agent, and the learned policy is deployed on all 12 tracks. The other 10 tracks and their expert demonstrations are not seen by the agent prior to evaluation, and so we use them to test the generalizability of the learned policy to unseen environments. In each episode, the car is spawned at a randomized initial position near the start of the track, so the agent cannot trivially solve the task by learning to drive straight without having to steer. The expert demonstrations consist of a single trajectory (egocentric, front-facing images) for all tracks provided by a human (the first author) controlling the demonstration vehicle with the objective of navigating from start to end of the tracks, driving straight, in the middle of the road, and avoiding collisions with obstacles such as parked cars.

We use the latent vector of the \textsc{rae} as the state representation, and the action space consists of change in steering and throttle values. Note that the expert demonstrations are required only during training to compute the reward. At test time, the agent imitates the expert without requiring access to expert demonstrations. 

We compare \voila{} against \gaifo{}, a state-of-the-art \ifo{} algorithm that does not explicitly seek to overcome viewpoint mismatch. While both \gaifo{} and \voila{} are \ifo{} algorithms that can imitate from video-only demonstration data, \gaifo{} has been evaluated predominantly in domains such as limbed-robot locomotion and manipulation, whereas \voila{} has been designed specifically for vehicle navigation domains. Additionally, \gaifo{} uses a learned reward function whereas in \voila{}, we propose a manually defined reward function that is not learned. 
To ensure a fair comparison, we provide each algorithm the same state representation, i.e., the latent code of the \textsc{rae}.
Further, since \gaifo{} is an on-policy algorithm whereas \voila{} relies on the off-policy \textsc{sac} algorithm, we allow \gaifo{} ten times more training timesteps than \voila{} (1 million vs. 100,000).
Finally, we report results for \gaifo{} using the policy that achieved maximum on-policy returns during training.

Fig. \ref{fig:bar_wo_vp} compares \voila{} and \gaifo{} without any viewpoint mismatch between the expert and imitator; we see that \gaifo{}, as expected, is able to imitate the expert demonstration on the training Tracks 1 and 2, but it fails to generalize to most unseen tracks. The policy trained with \voila{} performs better than \gaifo{} at imitating the expert demonstration on the training tracks, and also generalizes to unseen environments. Fig. \ref{fig:bar_w_vp} addresses both $Q_1$ and $Q_2$. In Fig. \ref{fig:bar_w_vp}, in the presence of viewpoint mismatch, we see that, also as expected, \gaifo{} is unable to imitate the expert on the training Track 2 and does not generalize to other environments. However, confirming our hypothesis, \voila{} is able to imitate the expert demonstration even in the presence of viewpoint mismatch on Tracks 1 and 2 and also generalizes to the other 10 unseen tracks. 
\gaifo{} performs best on Tracks 6 and 10 as shown in Fig. \ref{fig:bar_wo_vp}, where there are not many shadows or other distractions like parked cars, but fails to generalize to other unseen tracks.  




\subsection{Physical Experiments on the Jackal}
To answer $Q_1$ and $Q_2$ on a physical robot, we performed experiments using a Clearpath Jackal---a four-wheeled, differential drive ground robot equipped with a front facing camera. 
The environment we considered is an indoor office space, shown in Fig. \ref{fig:voila_trajs_w_mismatch}, which consists of carpeted floors, straight hallways, intersections, and turns.
There are also static obstacles such as benches, chairs, whiteboards, pillars along the wall, and trashcans, all of which the robot needs to avoid colliding with.

We evaluated \voila{} on a hallway patrol task, in which the Jackal robot begins at a start state (shown in Fig. \ref{fig:voila_trajs_w_mismatch}) and patrols around the building clockwise by taking the first right at intersections and driving straight in the hallways.
To obtain a video demonstration of this task from a physically different agent, a human (the first author) walked the patrol trajectory once while recording video using a mobile phone camera held approximately 4 feet above the ground (the robot's camera is at approximately 0.8 feet from the ground).
To contrast the imitation learning performance of \voila{} with and without any viewpoint mismatch, we perform additional experiments, henceforth called \voila{}-w/o-mismatch in which the expert demonstrations are collected onboard the deployment platform itself. These demonstrations are collected using the \textsc{ros} \texttt{move\_base} \cite{rosmovebase} navigation stack with a pre-built map of the environment and waypoints to patrol the environment while recording the egocentric visual observations from the front facing camera.

A \voila{} training episode consists of the robot starting at approximately the same start state (as shown in Fig. \ref{fig:voila_trajs_w_mismatch}) and exploring in the training environment until the agent reaches the \textit{done} state.
After each training episode, the robot was manually reset back to the start state by a human operator, and a new training episode began.
Training the navigation policy happened onboard the robot on a \textsc{gtx} 1050Ti \textsc{gpu}.

The \voila{} agent trained using demonstrations with and without viewpoint mismatch learned to imitate the expert within 60 minutes (100 episodes) and 90 minutes (120 episodes), respectively, of experiment time (including time taken to reset the robot at the end of an episode).
Fig. \ref{fig:voila_trajs_w_mismatch} shows the trajectory rollout (in green) of the policy learned using \voila{}, imitating the expert demonstration in the presence of viewpoint mismatch. Addressing $Q_1$, we see that \voila{} is able to successfully patrol the indoor environment in a real-world setting, as demonstrated by a physically different expert agent, in the presence of viewpoint mismatch.
 
 \begin{figure*}[!tb]
    \centering
    \includegraphics[scale=0.50]{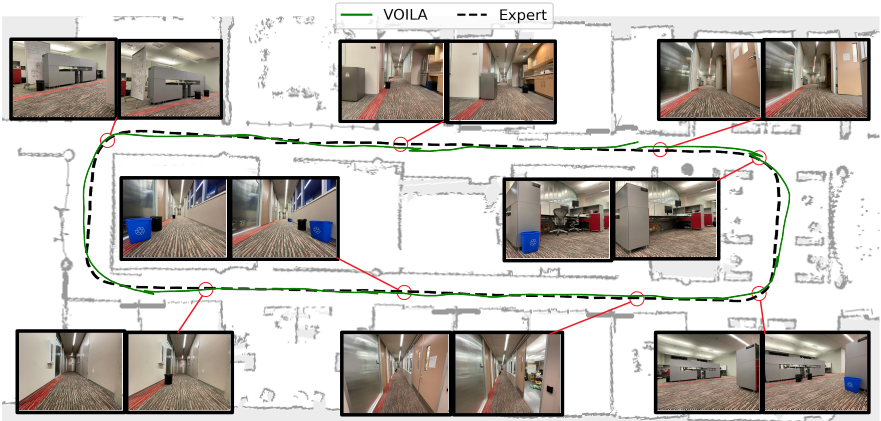}
    \caption{The \voila{} agent, trained in the unperturbed training environment, deployed here in the perturbed environment. We see that the learned policy is robust to the visual differences between the training and deployment environment, examples of which are provided as image pairs. The left image in each pair shows the training environment and the right image shows the perturbed environment.}
    \label{fig:voila_perturbed}
\end{figure*}

\begin{figure}[!tb]
    \centering
    \includegraphics[width=\columnwidth]{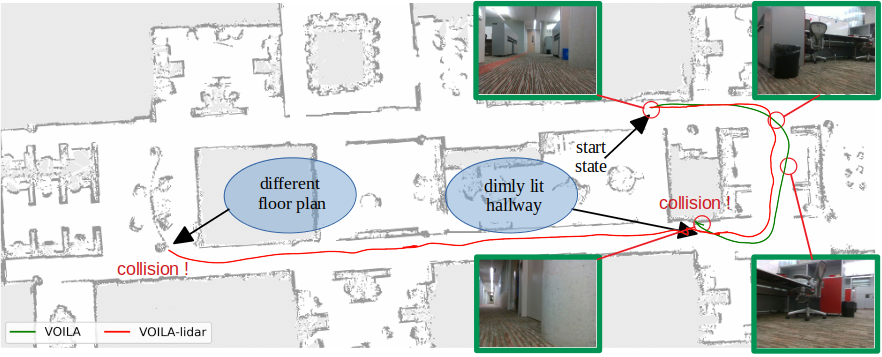}
    \caption{Deploying the policies learned using \voila{} in an environment with major visual (dimly-lit) and structural (floor plan) differences. While the agent succeeds for much of the trajectory, the \voila{} policy fails to fully generalize and patrol the hallway.}
    \label{fig:voila_4th_floor}
\end{figure}

To evaluate the generalizability of policies learned using \voila{} to unseen real-world conditions ($Q_2$), we deploy the policy learned by \voila{} in two new environments. First, we test generalizability of the learned policy in a `Perturbed Environment,' in which positions of movable objects such as trashcans, doors, whiteboards, chairs, and benches in the training environment on the same floor are perturbed as shown in Fig. \ref{fig:voila_perturbed}. We see that, with such environmental changes, \voila{} is able to successfully patrol the hallway without any collisions. Second, we deploy \voila{} on a different floor within the same building with major visual and structural changes from the training environment as shown in Fig. \ref{fig:voila_4th_floor}. While the robot succeeds for much of the trajectory, this experiment demonstrates the limitations of the current approach, as the robot collides with the walls in two places where there are large visual differences with the training environment.
The trajectory visualizations shown in Figures \ref{fig:voila_trajs_w_mismatch}, \ref{fig:voila_perturbed} and \ref{fig:voila_4th_floor} were generated using the \textsc{ros} \texttt{amcl} \cite{rosamcl} localization package for trajectories driven by the robot and using the SfM package \texttt{COLMAP} \cite{colmap} for the human demonstration.

In all experiments above, the end-to-end navigation policy takes as its input front camera images and predicts the actions of the agent. We performed an additional experiment to show that \voila{} can also learn a navigation policy over other sensor modalities, such as \textsc{lidar} range scans. To demonstrate this, we train \voila{} with \textsc{lidar} range scan as the policy's input, and observed that the agent learns to imitate the expert within 30 minutes of experiment time (3x faster compared to vision based navigation). As shown in Fig. \ref{fig:voila_trajs_w_mismatch}, in red, \voila{}-lidar shows the rollout trajectory of the policy learned using \voila{} with \textsc{lidar} range scans, demonstrating the successful imitation learning performance of \voila{}-lidar.

The Hausdorff distance between trajectories of the corresponding expert demonstration and the different policies learned using \voila{} are shown in Table \ref{table:1}. To provide context for the Hausdorff distance metric, we also show results for suboptimal and random trajectories. The suboptimal trajectory was collected by navigating along the hallway in a zig-zag route using \texttt{move\_base}, and the random trajectory was collected using a randomly initialized policy $\pi$, which fails quickly by crashing in the environment. We see that both \voila{} and \voila{}-w/o-mismatch perform well, and that the \textsc{lidar}-based policy provides better performance.


\begin{table}
\centering
\begin{tabular}{ ||c|c|c|| } 
 \hline
 Expert & Policy & Hausdorff Distance \\ 
 \hline \hline
  human demo & \voila{} &  \textbf{0.783} \\ 
  \hline \hline
 \texttt{move\_base} & \voila{}-lidar &  \textbf{0.487} \\ 
  \hline
 \texttt{move\_base} & \voila{}-w/o-mismatch &  0.665 \\ 
  \hline
 \texttt{-} & Suboptimal &  0.806 \\
  \hline
 \texttt{-} & Random &  1.192 \\ 
 \hline
\end{tabular}
\caption{Hausdorff distance between the expert trajectory and the policy rollout trajectory of \voila{}. Lower values indicate better imitation learning performance. Hausdorff distances for Suboptimal and Random policies are reported for additional context.}
\label{table:1}
\end{table}

\section{Conclusion and Future Work}
In this paper, we introduced Visual-Observation-only Imitation Learning for Autonomous navigation (\textsc{voila}), a new approach that enables imitation learning for autonomous robot navigation using a single, egocentric, video-only demonstration.
Furthermore, unlike prior methods, \textsc{voila} is robust to egocentric viewpoint mismatch.
\textsc{voila} formulates the imitation problem as one of reinforcement learning using a novel reward function that is based on keypoint matches between the expert and imitator's visual observations.
We showed through experiments, both in simulation and on a physical robot, that, by optimizing the proposed reward function using reinforcement learning, \voila{} could successfully find a good imitation policy that maps sensor observations directly to low level action commands.
We additionally performed experiments that tested the generalizability of policies trained using \voila{} to unseen environments.
One interesting direction for future work is to explore state representations that enable \voila{} to generalize better across environments.

\section{Acknowledgements}
\smaller{This work has taken place in the Learning Agents Research Group (LARG) at UT Austin.  LARG research is supported in part by NSF (CPS-1739964, IIS-1724157, NRI-1925082), ONR (N00014-18-2243), FLI (RFP2-000), ARL, DARPA, Lockheed Martin, GM, and Bosch.  Peter Stone serves as the Executive Director of Sony AI America and receives financial compensation for this work.  The terms of this arrangement have been reviewed and approved by the University of Texas at Austin in accordance with its policy on objectivity in research.}

\bibliographystyle{IEEEtran}
\bibliography{mybib}

\end{document}